\documentclass{article}
\pdfoutput=1
\usepackage{graphicx} % Required for inserting images
\usepackage{multicol}
\usepackage{indentfirst} 
\usepackage[margin=0.8in]{geometry}
\usepackage{amsmath}
\usepackage{caption}
\usepackage{url}
\usepackage[colorlinks,linkcolor=blue]{hyperref}
\usepackage{amssymb}
\usepackage{booktabs}

\usepackage{array} % 用于表格
\usepackage{float} % 用于 [H] 选项
\usepackage{graphicx} % 用于 \resizebox

\title{\textbf{Feature-Based Dual Visual Feature Extraction Model for Compound Multimodal Emotion Recognition} }
\date{} %

\author{
  \begin{tabular}{c}
    Ran Liu$^{1,2,3\ast}$,Fengyu Zhang$^{1,2\ast}$, Cong Yu$^{1,2\ast}$,Longjiang Yang$^{1,2\ast}$, Zhuofan Wen$^{1,2}$, Siyuan Zhang$^{1,2}$,\\  Hailiang Yao$^{1,2}$,Shun Chen$^{1,2}$,
     Zheng Lian$^{2}$, Bin Liu$^{1,2\dagger}$\\
    \multicolumn{1}{p{\textwidth}}{\centering\normalsize
    $^1$University of Chinese Academy of Sciences, China \\
    $^2$The State Key Laboratory of Multimodal Artificial Intelligence Systems, Institute of Automation, Chinese Academy of Sciences \\
    $^3$Tianjin Normal University\\
    \vspace{0.5em}
    $^\ast$These authors are equal contributors to this work. \\
    $^\dagger$Corresponding author
    }
  \end{tabular}
}

\begin{document}

\maketitle
\begin{multicols*}{2}

\section*{\centerline{Abstract}}
This article presents our results for the eighth Affective Behavior Analysis in-the-wild (ABAW\cite{kolliasadvancements} \cite{Kollias2025}) competition.Multimodal emotion recognition (ER) has important applications in affective computing and human-computer interaction. However, in the real world, compound emotion recognition faces greater issues of uncertainty and modal conflicts. For the Compound Expression (CE) Recognition Challenge,this paper proposes a multimodal emotion recognition method that fuses the features of Vision Transformer (ViT) and Residual Network (ResNet). We conducted experiments on the C-EXPR-DB and MELD datasets. The results show that in scenarios with complex visual and audio cues (such as C-EXPR-DB), the model that fuses the features of ViT and ResNet exhibits superior performance.Our code are avalible on\url{https://github.com/MyGitHub-ax/8th_ABAW}.
\section{Introduction}
\begin{figure*}
    \centering
    \includegraphics[width=1\linewidth]{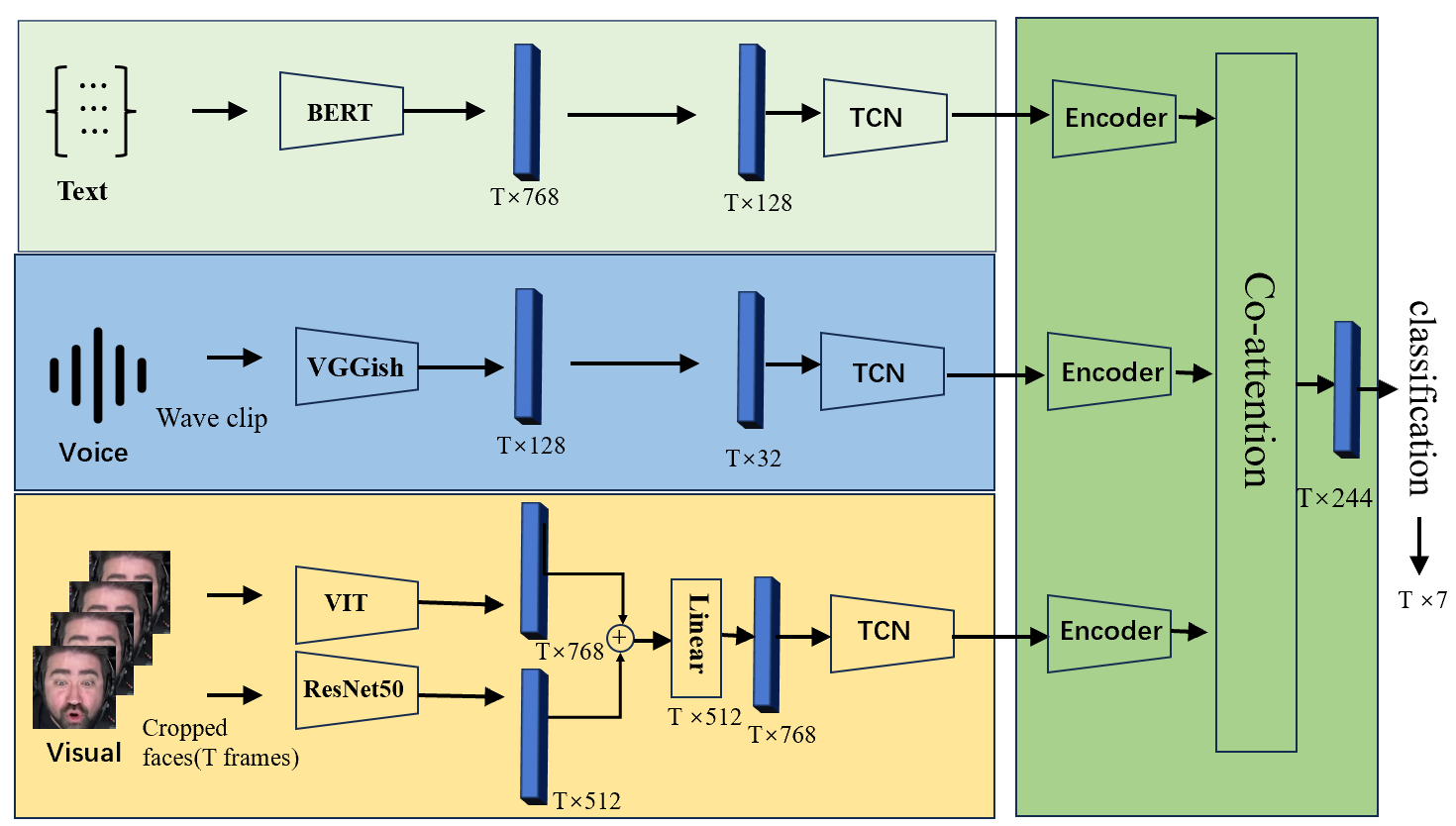}
    \caption{Feature-based Modeling.}
    \label{fig:enter-label}
\end{figure*}
Accurate emotion analysis is crucial for the development of human-centered technologies. In recent years, significant progress has been made in affective behavior analysis in real-world, unconstrained environments. However, traditional methods still face challenges in the in-the-wild context.
Multimodal emotion recognition (ER) has important applications in affective computing and human-computer interaction. However, in the real world, compound emotion recognition faces greater issues of uncertainty and modal conflicts. For the Compound Expression (CE) Recognition Challenge,this paper proposes a multimodal emotion recognition method that fuses the features of Vision Transformer (ViT) and Residual Network (ResNet). We conducted experiments on the C-EXPR-DB\cite{26} and MELD\cite{45} datasets.

\section{Method}
Emotion recognition has a wide range of applications in fields such as automatic behavior analysis, human-computer interaction, and health monitoring \cite{13}. Traditional research mainly focuses on basic emotions (such as anger, surprise, fear, etc.) \cite{4}. However, in real-world scenarios, human emotions are usually compound, that is, composed of multiple basic emotions \cite{kollias20247th,kollias2024distribution}. For example, "Fearfully Surprised" is a combination of surprise and fear. Nevertheless, the difficulty of compound emotion recognition lies in its cross-modal uncertainty and conflict \cite{23}. This paper proposes a multimodal method that combines the features of ViT and ResNet.
We propose a method to fuse the visual features of ViT and ResNet, which enhances the representational ability of visual features.This paper is based on the recent work on feature-based methods \cite{43}. It combines visual, audio, and textual information \cite{1}. In the visual modality, ResNet \cite{18} is used for feature extraction. For the audio modality, the VGGish \cite{19} model is employed to extract spectral features. In the textual modality, the encoding relies on BERT \cite{10}.
\begin{itemize}
   	\item We propose a method to fuse the visual features of ViT and ResNet, which enhances the representational ability of visual features.
	\item We conduct experiments on the C-EXPR-DB and MELD datasets, and systematically compare the feature-based and text-based methods.
	\item By integrating a multimodal fusion strategy, we improve the robustness of the model in the compound emotion recognition task.
\end{itemize}
\subsection{Dual Feature Extraction Module with Vision Transformer (ViT) and ResNet50}
In recent years, ViT \cite{vit} has achieved breakthroughs in image understanding tasks. Compared with traditional CNNs, ViT can better capture global information. However, in emotion recognition, directly using ViT may lead to the loss of local details. Therefore, combining ResNet for multi - scale feature fusion has become an effective strategy.
This paper utilizes ViT-Base and ResNet50 to extract visual features. ViT-Base extracts the feature \( F_{\text{vit}} \), where \( F_{\text{vit}} \in \mathbb{R}^{b \times 768} \), with \( b \) representing the batch size. ResNet50 extracts the feature \( F_{\text{resnet}} \), where \( F_{\text{resnet}} \in \mathbb{R}^{b \times 512} \). Finally, the two features are fused using concatenation:  

\[
F_v = \text{concat}(F_{\text{vit}}, F_{\text{resnet}})
\]

resulting in \( F_v \in \mathbb{R}^{b \times 1280} \).
Finally, a linear projection layer is applied to reduce the fused feature dimension to 512:  
\( F_v' = \text{Linear}(F_v) \), where \( F_v' \in \mathbb{R}^{T \times 512} \).
\subsection{Feature-based Modeling}
This paper follows the latest research \cite{43} to experiment with a feature-based approach for the Compound Emotion (CE) Recognition task (as shown in Fig. 1). Their method achieved excellent results in the recent ABAW challenge for the Compound Expression (CE) Recognition Challenge in videos. In our study, we modified the model to better suit our task.

Specifically, for the visual modality, we use ResNet50 \cite{18} and ViT \cite{vit}. The ResNet50 model is pre-trained on the MS-CELEB1M \cite{16} and FER+ \cite{13} datasets, while the ViT model is pre-trained on the ImageNet dataset. For the audio modality, we employ VGGish\cite{19}, and for the text modality, we utilize BERT \cite{10}. Additionally, we introduce a Temporal Convolutional Network (TCN) \cite{2} after each feature extractor to further exploit temporal information.

We adopt the co-attention module\cite{43} to attend to features from different modalities. This module constructs a single embedding per frame while leveraging a contextual window. The per-frame feature is then fed into a classifier head to predict emotions.
\subsection{Training setting}
We trained our model on the basic emotion dataset MELD and then tested it on the competition dataset C-EXPR-DB.

\section{Experiment}
\subsection{Dataset}
\begin{enumerate}
    \item C-EXPR-DB \cite{26}: It contains 56 compound emotion videos, covering 7 categories of compound emotions.
    \item MELD \cite{45}:This is a basic emotion dataset. The dataset is a multimodal emotion recognition dataset primarily used for affective computing and multimodal emotion analysis tasks. It is extracted from the dialogues of the TV series *Friends* and contains information from three modalities: visual, textual, and audio. The training set, validation set, and test set consist of 9,988, 1,108, and 2,610 utterances, respectively.
  
\end{enumerate}
\subsection{Evaluation metric}
For CE Challenge,We evaluate the performance of compound expressions recognition by the average F1 Score across all 7:
\begin{equation}
	\mathcal{P}_{CE}=\frac{\sum_{\mathrm{expr}} F_{1}^{\mathrm{expr}}}{7}
\end{equation}

According to the literature\cite{43}, we adopt the weighted F1 score as the evaluation metric for the MELD dataset. This metric can effectively mitigate the issue of class imbalance, similar to the approach used for the C-EXPR-DB dataset. It should be specifically noted that C-EXPR-DB follows the frame - level evaluation paradigm of C-EXPR-DB. In contrast, the MELD dataset only provides global video-level annotations (i.e., an entire video shares a single class label), so its evaluation needs to be conducted at the video level.

For the video-level prediction task, a three-stage post- processing strategy is used to integrate the frame-level prediction results:
\begin{itemize}
\item Majority Voting Mechanism: The distribution of predicted classes for all frames is counted, and the class with the highest frequency is selected as the final video label.

\item Logits Mean Aggregation: The mean of logits for all frames is calculated for each class to construct a video - level logits vector. The class corresponding to the maximum value is taken as the prediction result.

\item Probability Mean Aggregation: The principle is similar to logits aggregation, but the operation is performed on the normalized class probability values.

\item Frame - Level Ensemble Technique
In the ABAW CER competition, we utilize a time - window - based cross - model ensemble method. The specific process is as follows: First, obtain the independent prediction results of each model on a single frame. Then, for each target frame t, construct a sliding window that covers the current frame and its previous 9 frames (with a total length of 10). By counting the distribution of classes predicted by all models within the window, the majority voting mechanism is used to determine the final label of frame t. This strategy can effectively integrate the temporal context information and the complementary advantages of multiple models. 
\end{itemize}

Table 1 and Table 2 respectively show our results on the MELD and C-EXPR-DB datasets. 
\begin{center}
\centering

\begin{tabular}{cc}
\toprule
   \textbf{Prediction Method}  &\textbf{Feature-based}\\ \hline
 \textbf{Majority voting} & 56.94 \\ \hline
\textbf{Average logits} & 56.08 \\ \hline
\textbf{Average probabilities}  & 55.32 \\ \hline
 
\end{tabular}
\captionof{table}{Video-level weighted F1 score on MELD test set}
\label{tab:prediction_method}
\end{center}
\begin{center}
\centering

\begin{tabular}{cc}
\toprule
   \textbf{Prediction Method}  &\textbf{Feature-based}\\ \hline
 \textbf{Majority voting} & 60.34 \\ \hline
\textbf{Average logits} & 58.51 \\ \hline
\textbf{Average probabilities}  & 57.89 \\ \hline
 
\end{tabular}
\captionof{table}{Frame-level weighted F1 score on C-EXPR-DB test set}
\label{tab:prediction_method}
\end{center}

\section{Conclusion}
In this paper, we outline the approach we adopted for the Compound Expression (CE) Recognition Challenge of the 8th Asian Conference on Automatic Face and Gesture Recognition (ABAW). We have studied various pre-trained features from three common modalities: audio, visual, and text. We explored the application of pre-trained feature extraction. Specifically, we introduced a dual-feature extraction method that uses the Vision Transformer (ViT) and ResNet50 to extract visual features, employs BERT to extract text features, and utilizes VGGish to extract audio features. These features are then processed through a Temporal Convolutional Network (TCN) and fused at the feature level to obtain the final results.  
\bibliographystyle{unsrt}
\bibliography{references}
\end{multicols*}
\end{document}